\documentclass{article}

\PassOptionsToPackage{numbers, compress}{natbib}

\usepackage[preprint]{neurips_2023}

\usepackage[utf8]{inputenc} 
\usepackage[T1]{fontenc}    
\usepackage{hyperref}       
\usepackage{url}          
\usepackage{booktabs}     
\usepackage{amsfonts}     
\usepackage{nicefrac}     
\usepackage{microtype}      
\usepackage{xcolor}         
\usepackage{sidecap} 
\usepackage{graphicx} 
\usepackage{epsfig}
\usepackage{amsmath}
\usepackage{amssymb}
\usepackage{adjustbox}
\newcommand{\vv}[1]{\mathbf{#1}}
\usepackage{xcolor}
\newcommand\norm[1]{\left\lVert#1\right\rVert}
\usepackage[normalem]{ulem}
\newcommand{\linkk}[1]{\emph{\textcolor{magenta}{#1}}}
\definecolor{bright}{rgb}{0.8, 0.1, 0}

\newcommand\ourmethod{SDS-Complete}

\title{ Point-Cloud Completion with Pretrained Text-to-image Diffusion Models}

\author{%
  \hskip 0.6cm Yoni Kasten$^1$  \\
  \And
  \hskip 1cm Ohad Rahamim$^2$ \\
  \And
  \hskip 1cm Gal Chechik$^{1,2}$ 
  \AND
  \vspace{-25pt} \\
  $^1$NVIDIA Research \qquad
  $^2$Bar-Ilan University
}

\begin{document}
\maketitle

\begin{figure*}[h!]
    \centering
  \includegraphics[width=0.88\textwidth]{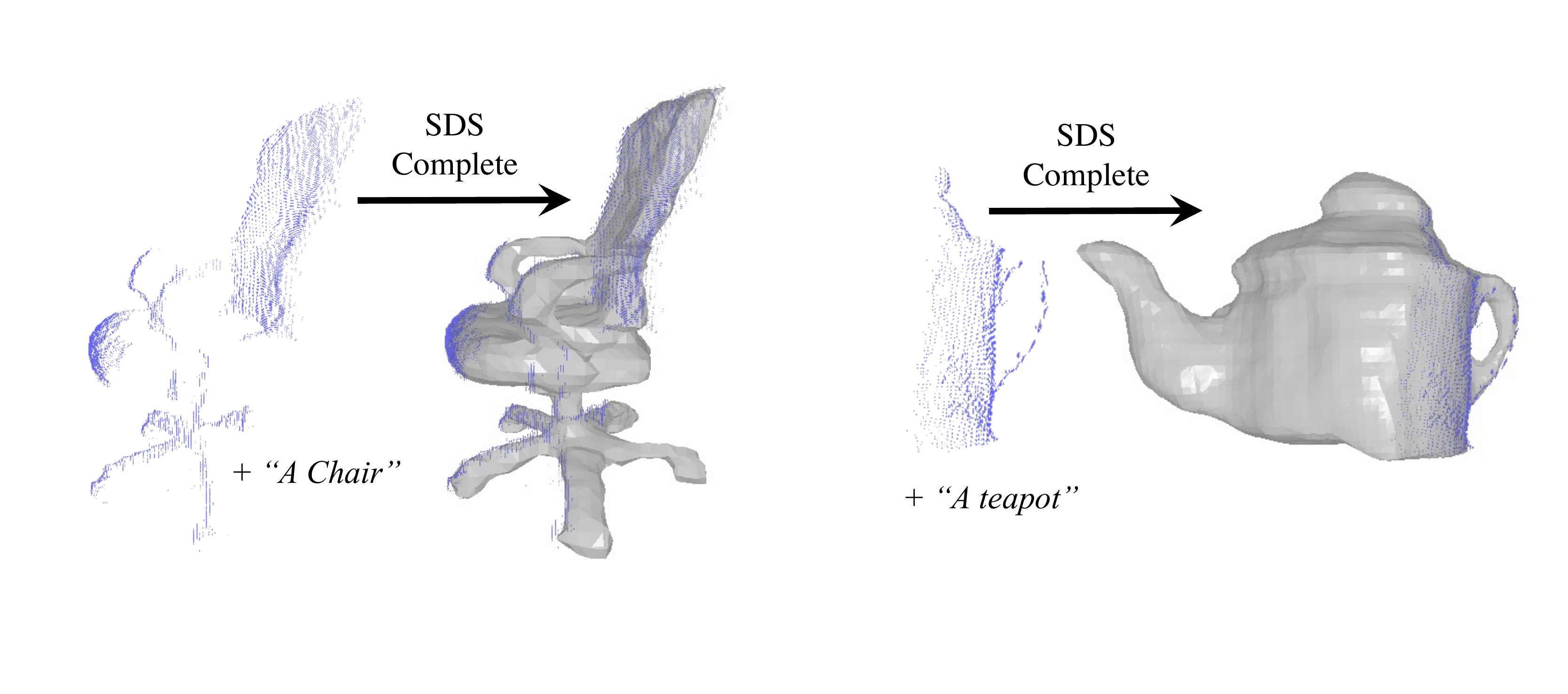}
  \caption{We present \textit{SDS-Complete}: A test-time optimization method for completing point clouds, captured by depth sensors, into complete surface representation using pre-trained text-to-image diffusion model. Our method takes two inputs: an incomplete point cloud (blue) and a textual description of the object ("A chair"). It  outputs a complete surface (gray) that is consistent with the input points (blue). 
  }
  \label{fig:teaser}
\end{figure*}
\begin{abstract}
Point-cloud data collected in real-world applications are often incomplete. Data is typically missing due to objects being observed from partial viewpoints, which only capture a specific perspective or angle. Additionally, data can be incomplete due to occlusion and low-resolution sampling.
Existing completion approaches rely on datasets of  predefined objects to guide the completion of noisy and incomplete, point clouds. However, these approaches perform poorly when tested on 
Out-Of-Distribution (OOD) objects, that are poorly represented in the training dataset.
Here we leverage  recent advances in text-guided image generation, which lead to major breakthroughs in text-guided shape generation. 
We describe an approach called \ourmethod{} that 
uses a pre-trained text-to-image diffusion model and leverages the text semantics of a given incomplete point cloud of an object, to obtain a complete surface representation.  %
\ourmethod{} can complete a variety of objects using test-time optimization without expensive collection of 3D information. We evaluate \ourmethod{} on incomplete scanned objects, captured by real-world depth sensors and LiDAR scanners. We find that it effectively reconstructs objects that are absent from common datasets, reducing Chamfer loss by 50\% on average compared with current methods.  Project page: \href{https://sds-complete.github.io/}{ \linkk{https://sds-complete.github.io/}}
\end{abstract}

\section{Introduction}
Depth cameras and LiDAR scanners enable us to capture the 3D geometrical structure of various objects in space. However, when used in the real world, various factors may significantly limit how well we can capture and reconstruct the full 3D geometry of objects from data alone. Specifically,  factors like 
self-occlusions, partial camera viewpoints, or limitations in sensor resolution may cause the scanner to capture incomplete or partially sampled 3D objects. To fully understand the three-dimensional world, one must address partial data and missing object parts.

Current approaches for point cloud completion demonstrate impressive results in handling in-domain shapes. However, due to the absence of large-scale datasets with a wide variety of shapes, these methods often face difficulties in dealing with shape classes that are outside their domain.
The need for extensive 3D data collection poses a significant challenge in developing a model capable of effectively completing diverse object classes encountered in real-world scenarios that require depth perception, such as indoor scene reconstruction or autonomous driving setups.

Surface completion has been extensively explored \cite{surveyComp}. Broadly speaking, some approaches focus on training models for specific object classes \cite{zhang2021unsupervised, wu2020multimodal,mescheder2019occupancy,peng2020convolutional} and some train class-agnostic models \cite{yan2022shapeformer, yu2021pointr,Williams_2022_CVPR}.
In general, existing methods achieve impressive results when tested on objects from the distribution they are trained on (in-domain). Unfortunately, their performance deteriorates dramatically for \textit{out-of-domain} objects, namely objects and classes that were not present in the training distribution.


In this work, we address this challenge of OOD objects by leveraging a pretrained text-to-image diffusion model.
It has been shown that these models, even though never trained on 3D data, can be used for 
text-guided 3D shape generation  \cite{poole2022dreamfusion}. 
This is done through the SDS loss, which measures the agreement of the 3D shape's rendered images with the model prior. 
Our key idea is that since text-to-image diffusion models were trained on a vast number of diverse objects, they contain a strong prior about the shape and texture of objects, and that prior can be used for completing missing parts. For example, given a partial point cloud, knowing that it corresponds to a chair can guide the completion process, because objects from this class are expected to exhibit some types of symmetries and parts.  

The key challenge in this approach is to combine the prior information from the diffusion model with the observed partial point cloud, to generate a complete shape that is faithful to the partial observations. We introduce \ourmethod{}: a point cloud completion method that uses the SDS-loss \cite{poole2022dreamfusion} to accurately complete object surfaces (Fig.~\ref{fig:teaser}, right) while being guided by input constraints of text and point clouds (Fig.~\ref{fig:teaser}, left).  To be consistent with the input points, we use a Signed Distance Function (SDF) surface representation \cite{Park_2019_CVPR,gropp2020implicit,Atzmon_2020_CVPR,VOLSDF}, and constrain the zero level set of the SDF to go through the input points. \ourmethod{} enables overcoming the limitations of working with OOD objects as it brings the semantics from a pretrained text-to-image diffusion model. That allows us to generate accurate and realistic 3D shapes from partial observations. 



We demonstrate that \ourmethod{}   generates completions for various objects with different shape types from two real-world datasets: the Redwood dataset \cite{choi2016large}, which contains various incomplete real-world depth camera scans, and the KITTI dataset \cite{behley2019iccv}, which contains object LiDAR scans from driving scenarios. In both cases, we outperform state-of-the-art methods for OOD objects, while showing comparable results on object classes that were used to train these methods.

In summary, this paper makes the following  contributions:  
(1) We formulate point cloud completion as a test-time optimization problem, avoiding collecting large datasets of 3D geometries for training. (2) We develop a new approach to combine the SDS loss \cite{poole2022dreamfusion}, with an empirical point-cloud, by using an SDF surface representation. 
    (3) We present a practical and unified approach for completing and preserving existing 3D content captured by different depth sensors (LiDAR or depth camera)  while  sampling realistic novel camera views for the SDS loss, that would complete the shape consistently.  
    (4) We demonstrate state-of-the-art completion results for objects considered to be out-of-distribution for point cloud completion.
    

\section{Related work}
\noindent\textbf{Surface Completion from Point Clouds.} Over the last years, neural network-driven approaches \cite{zhang2021unsupervised, wu2020multimodal, mescheder2019occupancy, peng2020convolutional} have demonstrated remarkable capabilities in reconstructing objects from incomplete or partial inputs. Early attempts with neural networks  \cite{choy20163dr2n2, dai2017shape, häne2017hierarchical, Stutz_2018} utilized voxel grid representations of 3D geometries due to their straightforward processing with off-the-shelf 3D convolutional layers. While voxels proved to be useful, they suffer from a space complexity issue, as their representation grows cubically. Consequently, these methods can only generate shapes with limited resolution. In contrast, point cloud representations \cite{fan2016point,achlioptas2018learning} have been leveraged to model higher-resolution geometries using neural networks. Several methods \cite{ xiang2021snowflakenet,yuan2018pcn} use such techniques for predicting the completed point cloud given a partial input point cloud. However, to obtain a surface representation, a surface reconstruction technique \cite{kazhdan2013screened} needs to be applied as a post-processing step, which can introduce additional errors.   
Recently, an alternative approach has emerged where the output surface is represented using neural representations \cite{mescheder2019occupancy,Park_2019_CVPR}.  
The advantage of these representations lies in their ability to represent the surface continuously without any discretization. \cite{mescheder2019occupancy,peng2020convolutional} trained deep neural networks and latent conditioned implicit neural representations on a dataset of predefined object classes \cite{chang2015shapenet}, to perform point cloud completion. While most deep methods for surface completion train a different model per object class, very recent methods have focused on training multi-class models, allowing for better generalization \cite{yan2022shapeformer, yu2021pointr}. PoinTr \cite{yu2021pointr} uses a transformer encoder-decoder architecture for translating a given input point cloud into a set of point proxies.  These point proxies are then converted into completed point clouds using FoldingNet \cite{yang2017foldingnet}. 
ShapeFormer \cite{yan2022shapeformer} directly reconstructs surfaces from incomplete point clouds using a transformer. 

Other recent works \cite{cheng2023sdfusion,li2023diffusion,autosdf2022} show progress in the task of shape completion given a partial surface, where \cite{autosdf2022} uses a transformer and autoregressive modeling, and  \cite{cheng2023sdfusion,li2023diffusion} employ diffusion processes that allow controlling the completion with text. However, these methods require a surface as input and cannot handle incomplete point clouds. Furthermore, their applicability is limited to the domain they are trained on.

In contrast to the above-mentioned methods, our method performs point cloud completion as a test-time optimization process using pre-trained available diffusion models, and therefore, we do not rely on any collection of 3D shapes for training, and we work on much broader domains.

\noindent\textbf{3D models from text using 2D supervision.} 
Several approaches used large vision-and-language models like CLIP \cite{radford2021learning} to analyze and synthesize 3D objects. 
Text2Mesh \cite{michel2021text2mesh}, CLIP-Mesh \cite{Mohammad_Khalid_2022} and DreamFields \cite{jain2022zero} present approaches for 
editing meshes, generating 3D models, and synthesizing NeRFs  \cite{mildenhall2020nerf}  respectively, based on input text prompts. The methods employ differentiable renderers to generate images while maximizing their similarity with the input text in CLIP space.

Diffusion models have recently gained attention for their ability to generate high-quality images \cite{yang2023diffusion}. One application of interest is textual-guided image generation  \cite{rombach2022high,NEURIPS2022_ec795aea}, where these models generate images based on text prompts, enabling control over the generated visual content. DreamFusion \cite{poole2022dreamfusion} pioneered the use of text-to-image diffusion models as guidance for text-guided 3D object generation. 
Latent-NeRF \cite{metzer2022latentnerf} enables the training of DreamFusion with higher-resolution images by optimizing the NeRF with diffusion model features instead of RGB colors. TEXTure \cite{richardson2023texture} and Text2Tex \cite{chen2023text2tex} use depth-aware text-to-image diffusion models to synthesize textures for meshes. Other recent works predict shapes directly from 2D images \cite{raj2023dreambooth3d,tang2023makeit3d,melaskyriazi2023realfusion}.  In contrast, our method uses the input text for completing partial point clouds, rather than editing or synthesizing 3D content. 

\section{Preliminaries}
\subsection{Volume Rendering} \label{sec::volume_rendering}
\paragraph{Neural Radiance Field}

A neural radiance field \cite{mildenhall2020nerf} is a pair of two functions:  $\vv{\sigma} : \mathbb{R}^3\rightarrow \mathbb{R}^{+}$ and $\vv{c}: (\mathbb{R}^3, \mathbb{S}^2) \rightarrow \mathbb{R}^3$, each represented by a Multilayer Perceptron  (MLP).  The function $\vv{\sigma}$ maps a 3D point $\vv{x} \in \mathbb{R}^3$ into a density value, and the function $\vv{c}$  maps a 3D point $\vv{x}$  and a view direction $\vv{v}\in \mathbb{S}^2$ into an RGB color. A neural radiance field can represent the geometric and appearance properties of a 3D object and is used as a differentiable renderer of 2D images from the 3D scene. Let $I$ be an image with a camera center $\vv{t} \in\mathbb{R}^3$, the pixel coordinate $\vv{u}=(u,v)^T \in \mathbb{R}^2$ is backprojected into a 3D ray $r_\vv{u}$, starting at $\vv{t}$ and going through the pixel $\vv{u}$ with a direction $\vv{v}\in \mathbb{S}^2$.  Let $\mu_1,\mu_2,\dots,\mu_{N_r}$ be sample distances from $\vv{t}$ on the ray $r_\vv{u}$, then the densities and colors of the radiance field are alpha composited from the camera center through the ray. The RGB image color $I(u,v)$ is calculated by: 
\begin{equation}\label{eq::volrender}
    I(u,v)=\sum_{i=1}^{N_r} w_i \vv{c}(\vv{t}+\mu_i\vv{v},\vv{v})
\end{equation}
where $w_i=\alpha_i \prod_{j<i} (1-\alpha_j)$ is the color contribution of the $i^{th}$ segment to the rendered pixel,  and $\alpha_i=1-\text{exp}\left(-\sigma(\vv{t}+\mu_i\vv{v})(\mu_{i+1}-\mu_i) \right)$  is the opacity of segment $i$.
Eq.~\eqref{eq::volrender} is differentiable with respect to the learned parameters of $\vv{c}$ and $\sigma$ and therefore, is used to train the neural radiance field. Let $\bar{I}$ be the ground truth image, then the MSE loss is used to train the neural radiance field: 
\begin{equation} \label{eq::mse}
    \mathcal{L}_{MSE} = \frac{1}{n} \sum_{i=1}^n \norm{I(\vv{u}_i)-\bar{I}(\vv{u}_i)}^2     
\end{equation}  
where $n$ is the number of pixels in the batch.

\paragraph{Volume Rendering of Neural Implicit Surfaces}
While the neural radiance field shows impressive performances in synthesizing novel views, extracting object geometries from a trained radiance field is not trivial. Defining the surface by simply thresholding the density $\sigma$ results in noisy and inaccurate geometry. We adopt the solution proposed by \cite{VOLSDF}. Let $\Omega\subset \mathbb{R}^3$ be the space occupied by the object, and $\mathcal{M}$ denotes the boundary of the surface. Then the SDF $f:\mathbb{R}^3 \rightarrow \mathbb{R}$ is defined by 
\begin{equation} 
    f(\vv{x})=(-1)^{\vv{1}_\Omega (\vv{x}) } \underset{\vv{y}\in \mathcal{M}}{ \text{min} } \  \norm{\vv{x}-\vv{y}}
    \label{eq: SDF}
\end{equation}
where $\vv{1}_\Omega (\vv{x})=\begin{cases} 1 \ \ \ \vv{x} \in \Omega \\ 0 \ \ \ \text{otherwise}\end{cases}$. Given $f$, the surface $\mathcal{M}$ is defined by its zero level set, i.e. \begin{equation} \label{eq::surface}
    \mathcal{M} = \{ \vv{x}\in \mathbb{R}^3 : f(\vv{x})=0\} 
\end{equation}
A signed distance function can be utilized for defining a neural radiance field density. Let $\vv{x}\in \mathbb{R}^3$ and $f:\mathbb{R}^3 \rightarrow \mathbb{R}$ be a 3D point and an SDF respectively, the density $\sigma(\vv{x})$ is defined by: \begin{equation} \label{eq::sdf_to_sigma}
    \sigma(\vv{x}) = \alpha \Psi_\beta (-f(\vv{x}))
\end{equation}  
where $ \Psi_\beta(s)$ is the Cumulative Distribution Function (CDF) of the Laplace distribution with zero mean and $\beta$ scale:
\begin{equation}
 \Psi_\beta(s)=\begin{cases}
     \frac{1}{2} \text{exp} \left(\frac{s}{\beta}\right) \ \ \ \ \ \ \ \ \ \ \ \ \ \  \   s \leq 0 \\
     1-\frac{1}{2} \text{exp} \left(-\frac{s}{\beta}\right) \ \  \ \ \  s > 0 
 \end{cases}
\end{equation}
and $\alpha$ and $\beta$ are parameters that can be learned during training (in our case, we set them to be constant, see details in the supplementary). It is then possible to train a neural radiance field, defined by the SDF $f$ and the neural color function $\vv{c}$, using the loss function defined by  Eq.~\eqref{eq::mse}.
\subsection{Score Distillation Sampling (SDS)}
\paragraph{Diffusion Models}
A diffusion model \cite{nichol2021improved,song2020denoising,rombach2022high} generates image samples from a Gaussian noise image, by inverting the process of gradually adding noise to an image. This process is defined as follows: at time $t=1,\dots,T$, a Gaussian noise $\epsilon \sim \mathcal{N}(\vv{0},I)$ is added to the image: $I_t=\sqrt{\bar{\alpha}_t}I+\sqrt{1-\bar{\alpha}_t}\epsilon$, where $\bar{\alpha}_t=\prod_{i=1}^t \alpha_i$,  $\alpha_t=1-\beta_t$ and $\beta_t \in (0,1)$ defines the amount of added noise. A denoising neural network $\hat{\epsilon}=\Phi(I_t ; t) $ is trained to predict the added noise $\hat{\epsilon}$ given the noisy image $I_t$ and the noise level $t$. The diffusion models are trained on large image collections $\mathcal{C}$ for minimizing the loss \begin{equation} \label{eq:diffusion_loss}
    \mathcal{L}_\text{D}=E\underset{I\in \mathcal{C}}{}\left[\norm{\Phi(\sqrt{\bar{\alpha}_t}I+\sqrt{1-\bar{\alpha}_t}\epsilon ; t)- \epsilon }^2 \right]
\end{equation}  
Given a pretrained $\Phi$, an image sample is generated by sampling a Gaussian noise image $I_T\sim \mathcal{N}(0,I)$ and gradually denoising it using $\Phi$. 

Diffusion models can be extended to be conditioned on additional inputs. Text-to-image diffusion models \cite{rombach2022high} condition $\Phi$ on a textual prompt embedding input $\vv{y}$, and train $\Phi(I_t ; t,y)$. Therefore, they can generate images given text and sampled Gaussian noise.
\begin{figure*}[t!]
\centering
  \includegraphics[width=\textwidth]{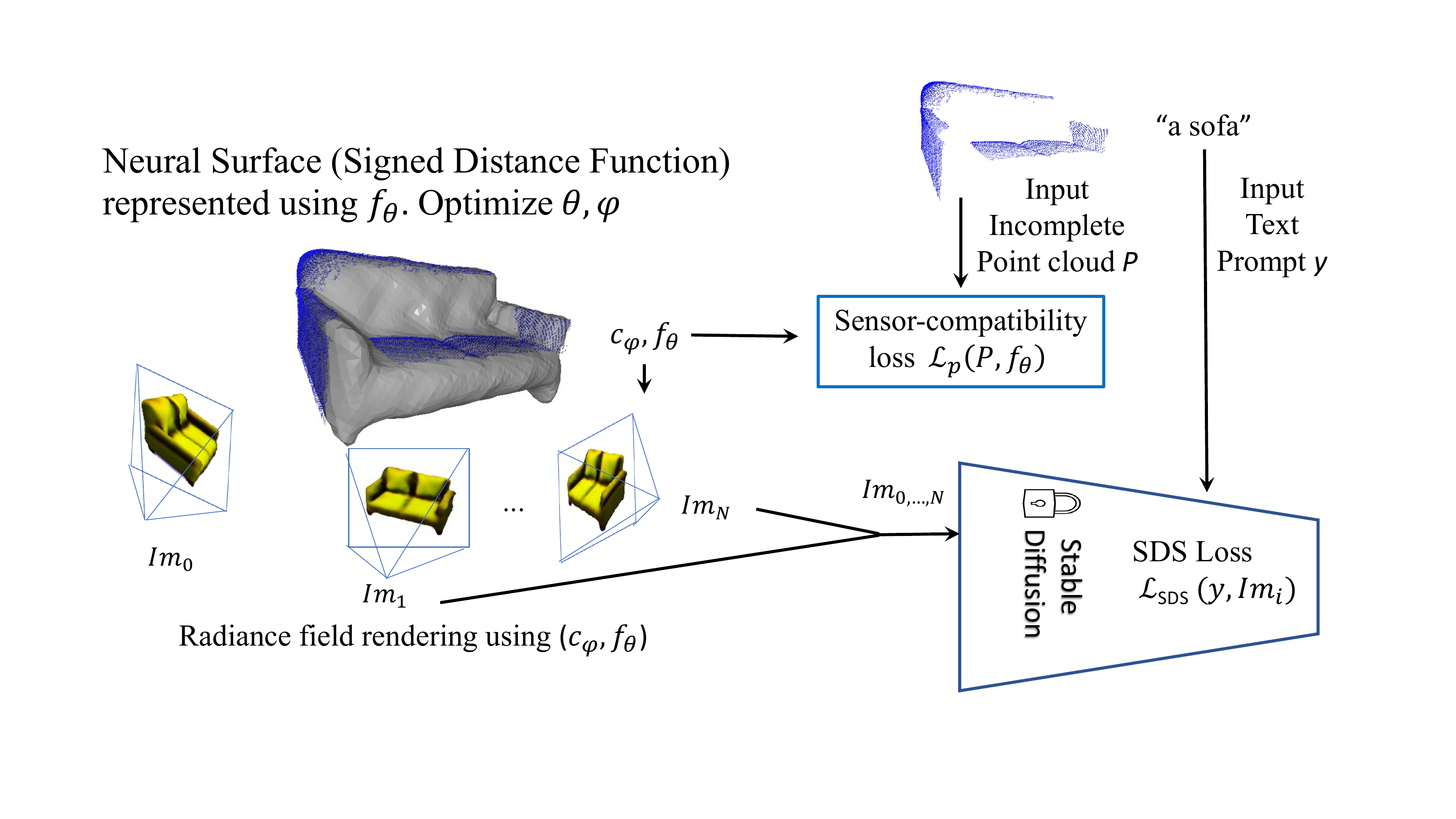}
  \caption{The components of our SDS-Complete approach. 
  Our method optimizes two neural functions: A signed distance function $f_\theta$ representing the surface and a volumetric coloring function $\vv{c}_\varphi$ (introduced in Sec. \ref{sec::volume_rendering}). Together, $(\vv{c}_\varphi,f_\theta)$ define a radiance field, which is used to render novel image views $Im_0,\ldots Im_n$. The SDS-Loss is applied to these renderings and encourages them to be compatible with the input text $\vv{y}$ (bottom left). To constrain the surface to lie on the input points, we encourage the signed distance function to be zero at the input points (Sensor compatibility loss). 
}
  \label{fig:pipeline}
\end{figure*}

DreamFusion\cite{poole2022dreamfusion} uses a pretrained, and fixed, text condition diffusion model $\Phi(I_t ; t,y)$ and uses it to train a NeRF model from scratch,  given a textural description embedding $\vv{y}_0$. In each iteration, a camera view is sampled and used to render an image $I_0$ from the  NeRF model. $I_0$ is differentiable with respect to the learned parameters of the NeRF model ($\theta_{\text{NeRF}}$), and used as an input to $\Phi(I_0 ; t,y)$. The Score Distillation Sampling (SDS) loss is then applied:
\begin{align}\label{eq::sds}
     \nabla_{\theta_{\text{NeRF}}}\mathcal{L}_{\text{SDS}}(I_0)=E_{t,\epsilon}   \left[(w(t)\Phi(\sqrt{\bar{\alpha}_t}I_0+\sqrt{1-\bar{\alpha}_t}\epsilon ; t,y_0)- \epsilon) \nabla_{\theta_{\text{NeRF}}} I_0\right]
\end{align}
Note that $ \nabla_{\theta_{\text{NeRF}}}\mathcal{L}_{\text{SDS}}$ is the gradient with respect to $\theta_{\text{NeRF}}$ of Eq.~\eqref{eq:diffusion_loss}, where the Jacobian of $\Phi$ is omitted for stability and efficiency. Intuitively, if $I_0$ looks like a natural image, and is compatible with $\vv{y}_0$, then the pretrained diffusion model predicts the added noise successfully, resulting in low values for $\mathcal{L}_\text{D}$. By updating the NeRF's weights according to  Eq.~\eqref{eq::sds},  $\mathcal{L}_\text{D}$ is reduced, and as a result, the rendered images become more compatible with $\vv{y}_0$.

\section{Our Method}

\textbf{Inputs and components of our system.}
The overall scheme for our method is depicted in Fig.~\ref{fig:pipeline}. We address the problem of completing a surface given incomplete point cloud measurements captured by a depth sensor.
These input measurements  (Fig.~\ref{fig:pipeline}  top-right), include a set of 3D input points $P=\{\vv{p}_1,\vv{p}_2,\ldots,\vv{p}_N\}$ and a text description embedding $\vv{y}$ of the incomplete object. We assume that $P$ is captured by a depth sensor like a depth camera or a LiDAR sensor, and that the internal parameters of the sensor are known. We further assume that the point cloud is segmented out from the original scan, namely, that all the points in $P$ belong to a single object that is described by $\vv{y}$. 
A sensor ray $i$ is associated with a binary value mask $M_i\in \{0,1\}$, indicating whether this ray intersects the surface at a point that belongs to $P$. The ray $i$ is also associated with the ray's distance from the sensor to the surface $D_i\in \mathbb{R}$ if $M_i=1$. 
Lastly, for our camera sampling process (see Sec.\ref{sec::camera}), we assume that the original, non-segmented scan, contains points from the world's ground plane, that are used to estimate the plane's parameters $\vv{l}\in \mathbb{P}^3$  \cite{hartley2003multiple}.

\subsection{Loss Terms}
Our method optimizes for the complete object surface represented by a neural signed distance function $f_\theta:\mathbb{R}^3 \rightarrow \mathbb{R}$, (see Eq. ~\ref{eq: SDF}), and a neural color function  $\vv{c}_\varphi: \mathbb{R}^3 \rightarrow \mathbb{R}^3$, where $\theta$ and $\varphi$ represent the learned parameters of the neural functions.  
As described in Section~\ref{sec::volume_rendering},  these two functions form a neural radiance field and can be optimized using the rendered images of the 3D volumetric functions. In contrast to \cite{poole2022dreamfusion}, the object surface is defined directly by $f_\theta$, as its zero level set (Eq.~\eqref{eq::surface}). To constrain the surface to go through the input points we encourage the signed distance to be zero at these points (Fig.~\ref{fig:pipeline}, middle-right), using the following point-cloud loss: 
\begin{equation}
    \mathcal{L}_p=\frac{1}{N}\sum_{i=1}^N | f_\theta\left(\vv{p}_i\right) |.
\end{equation}  
At each iteration, we render the radiance field from the sensor perspective. Each rendered pixel $i$ is associated with its expected rendered opacity and distance from the surface, denoted by  $\tilde{M}_i$ and $\tilde{D}_i$ respectively. We use the input opacities and distances  to  constrain the optimized surface to match the mask and depth sensor observations: 
\begin{equation}
    \mathcal{L}_\text{m}=\frac{1}{K}\sum_{i=1}^K  |M_i - \tilde{M}_i | \quad, \quad  
    \mathcal{L}_\text{d}=\frac{1}{K}\sum_{i=1}^K  \norm{D_i - \tilde{D}_i }^2  \quad, 
\end{equation}  
where $K$ is the number of sensor rays. 
To constrain $f_\theta$ to form a valid SDF, we apply the Eikonal loss regularization introduced in \cite{gropp2020implicit}:
\begin{equation}
    \mathcal{L}_{eikonal}=\frac{1}{|P_{eik}|}  \sum_{\vv{p} \in P_{eik} } |\norm{ \nabla f_\theta\left(\vv{p}_i\right) } -1| \quad ,
\end{equation}
where $P_{eik}$ contains both $P$ and uniformly sampled points from the region of interest.

While $\mathcal{L}_{m},\mathcal{L}_d,\mathcal{L}_p$ and $\mathcal{L}_{eikonal}$, constrain the optimized surface to match the information that is captured by the sensor, the losses do not provide any signal for the occluded missing content that cannot be captured by the depth sensor. A semantic prior is required in order to complete the unobserved part of the surface. For that, we utilize the input text embedding $\vv{y}$ and a pretrained text-to-image diffusion model $\Phi$. Our goal is to use $\Phi$ to supply the semantic prior for the unobserved parts, such that any rendered image of the object would be compatible with $\vv{y}$. To this end, we render random object views using our radiance field and apply the SDS loss (Eq. \eqref{eq::sds}) with the input text embedding $\vv{y}$ to optimize $f_\theta$ and $\vv{c_\varphi}$ (Fig.~\ref{fig:pipeline}, bottom-right).

Finally, we use the known world plane to further regularize the surface from drifting below the ground: 
\begin{equation}
    \mathcal{L}_{plane}= \sum_{\vv{p} \in P_{\text{uniform}} } \text{max}\left(-f_\theta(\vv{p}),0\right),
\end{equation}
where $P_{\text{uniform}}$ is a set of uniformly sampled $3D$ points below the plane in the region of interest.
Our total loss is:
\begin{equation}
\mathcal{L}_{\text{total}}=\delta_{m}\mathcal{L}_{m}+\delta_{d}\mathcal{L}_d+\delta_{p}\mathcal{L}_p +\delta_{eikonal}\mathcal{L}_{eikonal}+\delta_{plane}\mathcal{L}_{plane}+\mathcal{L}_{\text{SDS}},
\end{equation}
where $\delta_{m},\delta_{d},\delta_{p},\delta_{eikonal}$ and $\delta_{plane}$ are the coefficients that define the weights of the different loss terms relative to the SDS loss.
\subsection{Camera handling}
\label{sec::camera}
To keep the generated content consistent with the existing partially observed object, careful handling of camera sampling is needed. In contrast to \cite{poole2022dreamfusion} where the SDS loss is used to generate a 3D object ``from scratch", in our case sampling camera positions uniformly at random, results in inferior results (see ablation study in Fig. \ref{ablation}).  

Instead, we developed a ``curriculum" for sampling camera poses. Let $C_0 = (R_0,\vv{t}_0)$ be the original camera-to-world pose of the depth sensor. To preserve the roll angle of $C_0$ with respect to the object and prevent rendering flipped or unrealistically rotated images, we define the azimuth and elevation deviation from $C_0$ with respect to the segmented world plane. Specifically, let $\vv{n}_\vv{l}\in\mathbb{S}^2$ be the normal to the plane $\vv{l}$, we define the azimuth rotation update to be $R_{\text{azimuth}} = $ $\mathcal{R}(\vv{n}_\vv{l},\gamma_\text{azimuth})$, where $\mathcal{R}(\vv{n},\gamma) $ is the Rodrigues' rotation formula for a rotation around the unit vector $\vv{n}$, with $\gamma$ degrees. 
Similarly, let $\vv{a}_0$ be the normalized  principal axis direction of $C_0$, we define the elevation rotation update by $   R_{\text{elevation}}=\mathcal{R}(\vv{n}_\vv{l}\times \vv{a}_0, \gamma_\text{elevation})$. Assuming that the origin is located at the object's center, an updated camera, $C_{\text{update}}$, for $\gamma_\text{azimuth}$ and $\gamma_\text{elevation}$ degrees, is given by: 
\begin{equation}
    C_{\text{update}}=(R_{\text{azimuth}}R_{\text{elevation}}R_0, R_{\text{azimuth}}R_{\text{elevation}}\vv{t}_0).
\end{equation}
During training, we start by applying the SDS loss on the rendered image from $C_0$ pose, and then we gradually increase the sampling range of the deviation angles until the entire object is covered. By initially applying the SDS loss on images rendered from the depth sensor's perspective, the colors of the observed part of the object are optimized first to be consistent with $\vv{y}$, and then, when the sampling range increases, the rest of the object's colors and geometry are completed accordingly.       



\begin{figure}[t]
\textbf{\hspace{10pt} GT \hspace{35pt} Ours \hspace{20pt} ShapeFormer \hspace{20pt} Sinv \hspace{25pt} cGAN \hspace{30pt} PoinTr \hspace{25pt} Input}
\hrule
        \includegraphics[scale=0.09]{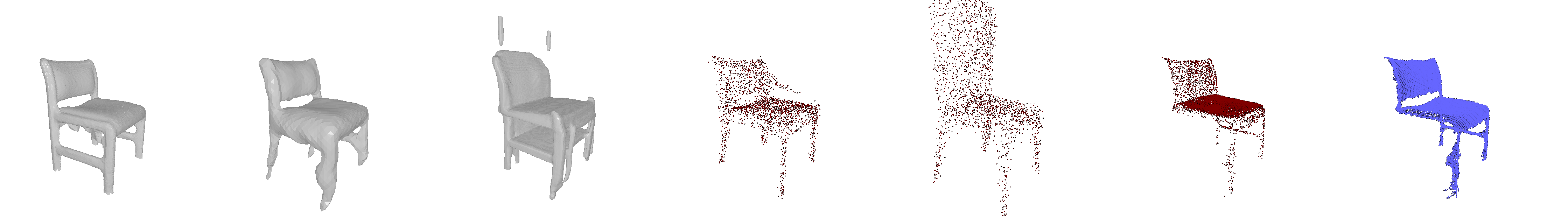}
        \includegraphics[scale=0.09]{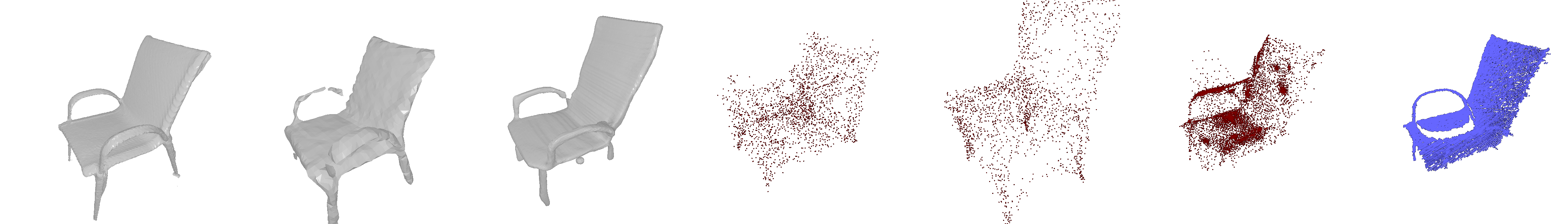}
        \includegraphics[scale=0.09]{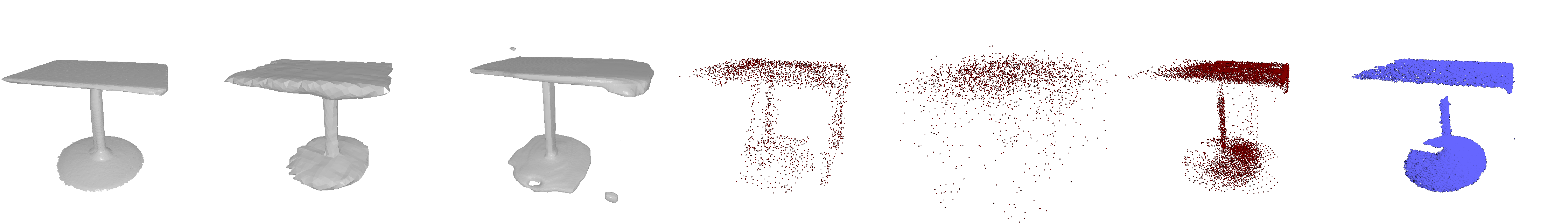}
    \caption{\textbf{In-domain completion. } Comparison with traditional shape completion methods on the Redwood dataset for in-domain objects. 
    Red represents methods that output point-clouds. \ourmethod{} completes unseen parts with accuracy that is comparable to previous work.}
    \label{fig:comperison figures}
\end{figure}

\section{Experiments}

We conduct an evaluation of our model, on two real-world datasets that encompass a diverse array of general objects. Our primary objective in selecting these datasets is to demonstrate that our proposed method can handle diverse variations of object types that are not confined to any specific domain. 

\begin{figure}[ht]
    \centering
     \includegraphics[width=0.85\textwidth]{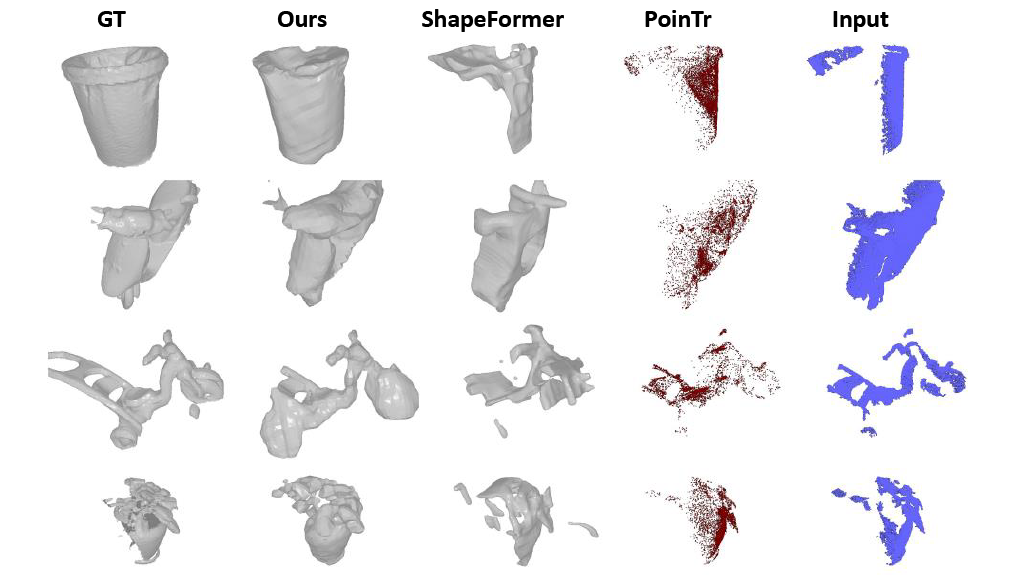}
        \caption{\textbf{Out-of-domain comparisons on the Redwood dataset. } A qualitative comparison between \ourmethod{} to multi-class methods that aim to exhibit generalization capabilities. Notably, \ourmethod{} produces more accurate completions.}
    \label{fig:OOD comperison figures}
\end{figure}

\paragraph{Datasets.}
We assessed the performance of our model by utilizing partial point clouds obtained from depth images and LiDAR scans. The Redwood dataset \cite{choi2016large} comprises a diverse collection of object scans.  On Tab. \ref{redwood-table} we compare the Chamfer distance in mm, for objects with existing $360^{\circ}$ scans that we refer to as a GT. We further tested our model on the KITTI LiDAR dataset \cite{behley2021ijrr, geiger2012cvpr}, which contains  incomplete point clouds of objects in real-world scenes captured by LiDAR sensors.  In contrast to the Redwood dataset, for KITTI,  there is no ground-truth data available for quantitative evaluation, and therefore, we only compare the outputs qualitatively. We note that there exist large datasets for 3D car shapes e.g. ShapeNetCars \cite{yuan2018pcn}  which were used for training the baseline methods and therefore we consider cars as in-domain objects.  To evaluate our performance on OOD classes compared to the baseline methods, we further include other object classes such as trucks and motorcycles.  
\paragraph{Baselines}
We conducted a comparative analysis of \ourmethod{} in comparison to several other point-cloud completion approaches, including PoinTr \cite{yu2021pointr} and ShapeFormer \cite{yan2022shapeformer}. Both PoinTr and ShapeFormer are characterized by their adoption of multi-model training methodologies. Furthermore, we extended our analysis to include methods that specialize in specific object-class: cGAN \cite{wu2020multimodal} and shape-inversion \cite{zhang2021unsupervised} have per-class trained models for chairs and tables. During the inference stage, object alignment was executed by utilizing the world's ground plane, denoted as $\vv{l}\in \mathbb{P}^3$ \cite{hartley2003multiple}, to employ the same alignment procedure that these methods applied during their training phase.

\paragraph{Results on Redwood}
We measure the Chamfer distances to quantify the dissimilarity between the generated completions and their corresponding ground-truth shapes. Our evaluation encompasses two distinct groups of objects: one contains objects from the same object classes of the baselines' training data (\cite{chang2015shapenet}), whereas the other group contains OOD objects that lack comprehensive pre-existing training data. The results are presented in Tab.~\ref{redwood-table} and demonstrate that \ourmethod{} gets state-of-the-art results on OOD objects while remaining comparable to the baselines on in-domain objects. We further show qualitative results on in-domain and out-domain objects in Fig. \ref{fig:comperison figures} and \ref{fig:OOD comperison figures} respectively. 
It can be observed that our method has the capability to maintain consistent performance across both in-distribution and OOD objects, while the completions generated by other methods for OOD objects exhibit unpredictability and deviate from the intended shapes, leading to inferior performance. 

To demonstrate the importance of each component of our method we present an ablation study in Fig. \ref{ablation}. As can be seen,  without the SDS-loss, our model has no understanding of object characteristics like the fact that the chair has four legs and a straight back-side. Without the SDF representation, it is not possible to apply the point cloud constraints directly on the surface which results in an inferior ability to follow the partial input. Lastly, it can be seen that our camera sampling ``curriculum" improves the completion compared to a random camera sampling, by preserving the consistency of the generated content with the existing sensor measurements.

\begin{figure}
    \centering
    \includegraphics[width=0.85\textwidth]{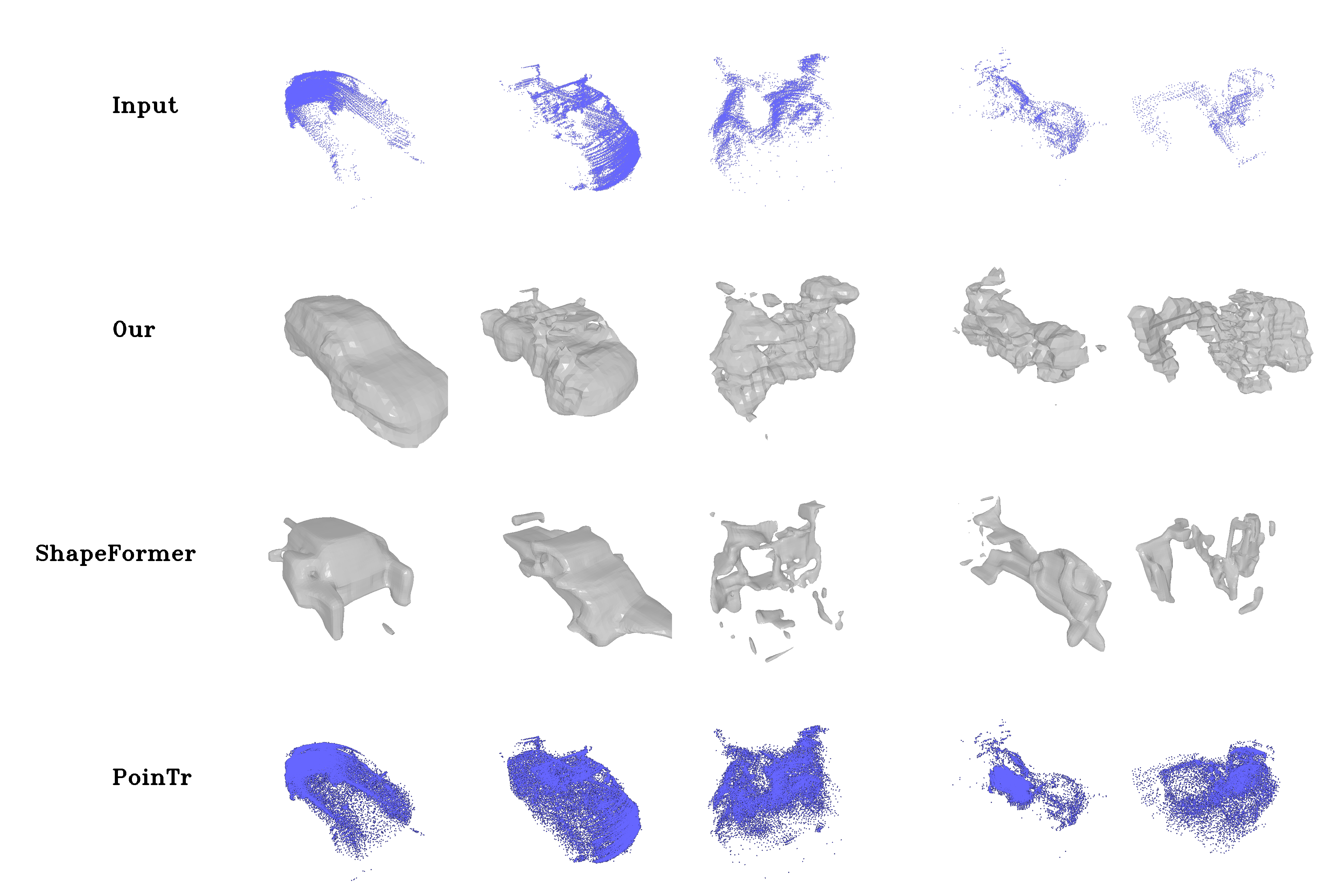}
    \caption{Completion results on the KITTI dataset. A qualitative comparison with previous methods that are trained on a dataset containing car shapes. Notably, \ourmethod{}, produces results that better complete the shape. Other methods fail to produce meaningful shapes for OOD objects}
    \label{fig:kiti}
\end{figure}

\begin{figure}
\centering
    \begin{tabular}{c} \includegraphics[width=0.65\textwidth]{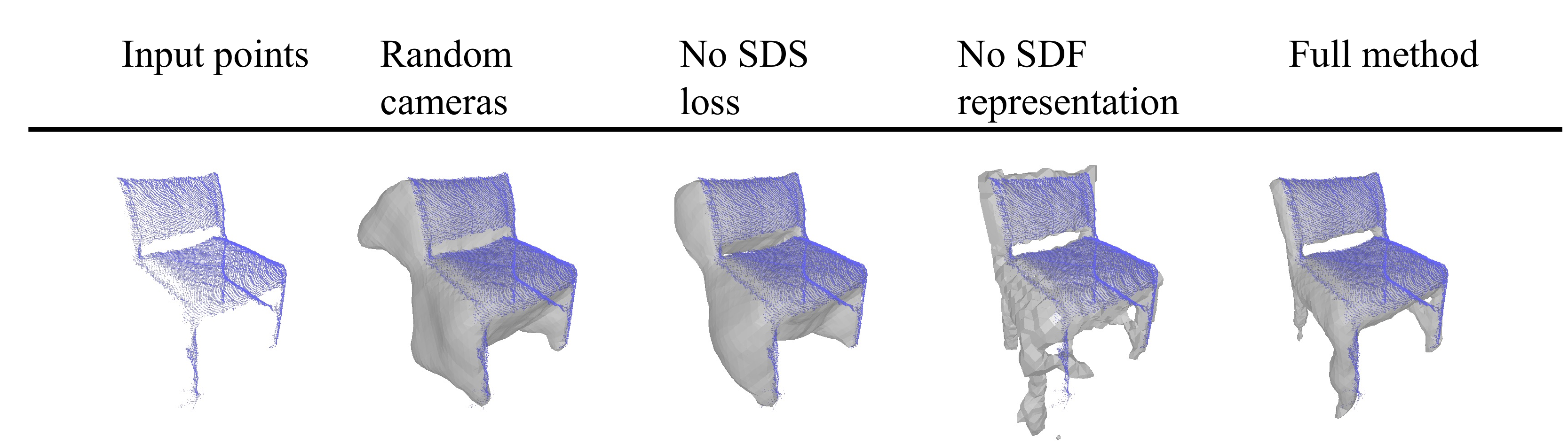}\\
     \label{fig:ablation}
    
    \hspace{-55pt}
    \scriptsize
    \begin{tabular}{p{0.17\textwidth}p{0.1\textwidth}p{0.1\textwidth}p{0.1\textwidth}p{0.1\textwidth}p{0.1\textwidth}}
    Average Chamfer $\downarrow$& &  43.5 & 43.8 & 59.2 & \textbf{30.5} \\
    \bottomrule
    \end{tabular}
    \label{tab:ablation}
    \end{tabular}
\caption{An ablation study for demonstrating the contribution of each part of our method. Random cameras: running without our camera handling that is described in Sec.~\ref{sec::camera}. No SDS
loss: using all losses but the SDS loss. No SDF representation: running with a density function as in \cite{mildenhall2020nerf}.  Below, we compare the average Chamfer distance over the evaluated 10 Redwood scans.}
   \label{ablation}
\end{figure}

\paragraph{Results on KITTI}
We present qualitative comparisons  in Fig. \ref{fig:kiti}. Besides our method, we include ShapeFormer \cite{yan2022shapeformer} and PoinTr \cite{yu2021pointr}, both of which were trained on ShapeNetCars \cite{yuan2018pcn}. Notably, \ourmethod{} exhibited better completion results, particularly when confronted with objects with no trainable data. In the supplementary, we further present a user-study evaluation to compare the performance of the 3 methods.

\begin{SCtable}
    \scalebox{1.01}{
    \setlength{\tabcolsep}{3pt} %
        \begin{tabular}{l rrrrr}
         \toprule
         \midrule
         Object & Shape  & PoinTr & cGAN & Sinv & \ourmethod{} \\
          &  Former &  &  &  &  (ours) \\
         \midrule
         old chair$\dagger$ & 23.2 & 34.1 & 33.2 & 36.7 & \textbf{19.3} \\
         outside chair$\dagger$ & 25.9 & 29.6 & 42.8 & 28.7 & \textbf{22.6} \\
         one lag table$\dagger$ & 39.7 & 21.6 & 99.4 & 24.9 & \textbf{20.3} \\
         executive chair$\dagger$ & 33.6 & 43.9 & 208 & \textbf{20.6} & 23.7 \\
         \quad Average (in) & 30.6  & 32.3 & 95.8 & 27.7 & \textbf{21.5} \\
         \midrule
         trash can & 136.4 & 137 & - & - & \textbf{36.4} \\
         plant in vase & 60.8 & 41 & - & - & \textbf{29.5} \\
         vespa & 79.4 & 70.3 & - & - & \textbf{57.6} \\
         tricycle & 65.2 & 60.4 & - & - & \textbf{39} \\
         couch & 43.9 & 87.4 & - & - & \textbf{36.5} \\
         office trash & 68.8 & 49.7 & - & - & \textbf{20.5} \\
         \quad Average (out) & 75.7 & 74.3 & - & - & \textbf{36.6} \\
         \midrule
         \quad Average & 60.4 & 59 & - & - & \textbf{30.5} \\
         \midrule
        \end{tabular}
    }
    \caption{\textbf{Chamfer loss (lower is better) for objects from the Redwood dataset}. 
    $\dagger$ represents in-domain objects. cGAN and Sinv models solely focused on chairs and tables.
    ``Average" denotes the mean performance on all 10 objects, ``Average (in)" refers to in-domain objects, and ``Average (out)" refers to OOD objects.}
    \label{redwood-table}
\end{SCtable}

\paragraph{Limitations} The major factor limiting our method is the application of the SDS loss with low resolution images due to GPU memory limitation, which requires a lot of sampling views until the object is completed. Our SDF model is initialized to a sphere and therefore cannot handle well objects with components that have disc topology. In the supplementary, we show failure cases of our method and list additional implementation details. 
\section{Conclusions}

We presented \ourmethod{}, a novel test time optimization approach for 3D completion that leverages a text-to-2D pre-trained model to enable the reconstruction of a wide variety of objects. To adapt the SDS-loss for the use of point clouds, we incorporated an SDF representation and constrained the surface to lie on the input points. We successfully applied the SDS-loss on images rendered from novel views and completed the missing part of the object by aligning the images with an input textual description. By handling the camera sampling carefully we maintained the consistency of the completed part with the input captured part. This enabled us to produce superior results even on previously unconsidered objects for completion.  In the future, we would like to utilize advances in text-to-3D for achieving higher-quality completions.
\section{Acknowledgments}
We thank Lior Yariv, Dolev Ofri, Or Perel and Haggai Maron for their insightful comments. We thank Lior Bracha and Chen Tessler for helping with the user study. This work was funded by a grant to GC from the Israel
Science Foundation (ISF 737/2018), and by an equipment
grant to GC and Bar-Ilan University from the Israel Science Foundation (ISF 2332/18). OR is supported by a PhD
fellowship from Bar-Ilan data science institute (BIU DSI).

{\small
\bibliographystyle{abbrv}
\bibliography{egbib}
}

\appendix

\section{Sensitivity to Textual Description}

\subsection{Completion with Different Text Descriptions}
Our approach operates by combining a partial input point cloud with a text description that guides the model when completing missing parts of the object. 

\begin{figure}[h]
    \centering
    \includegraphics[width=0.95\textwidth]{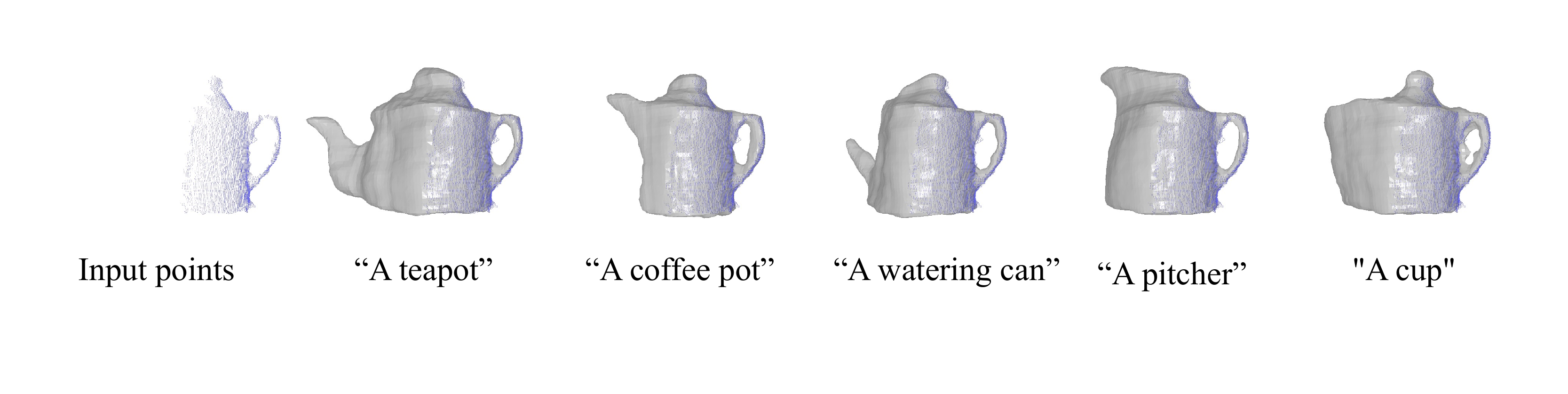}
    \vspace{-30pt}
    \caption{Completion of the same input, with different text descriptions. Results obtained with our method for the partial point cloud of scan "08754". While the handle and the top part of the object are constrained by the input point cloud, the model completes the other side of the object according to the input text. \label{fig:fig_control_text}}
\end{figure}

We tested the effect of changing the text prompt while keeping the same input point cloud. Fig.~\ref{fig:fig_control_text} shows results for completing Redwood's scan "08754" of a partially captured teapot (Fig.1 in the main paper). Completing the point cloud with other text descriptions demonstrates how the text controls the shape.  

\begin{figure}[h]
    \centering
    \includegraphics[width=0.95\textwidth]{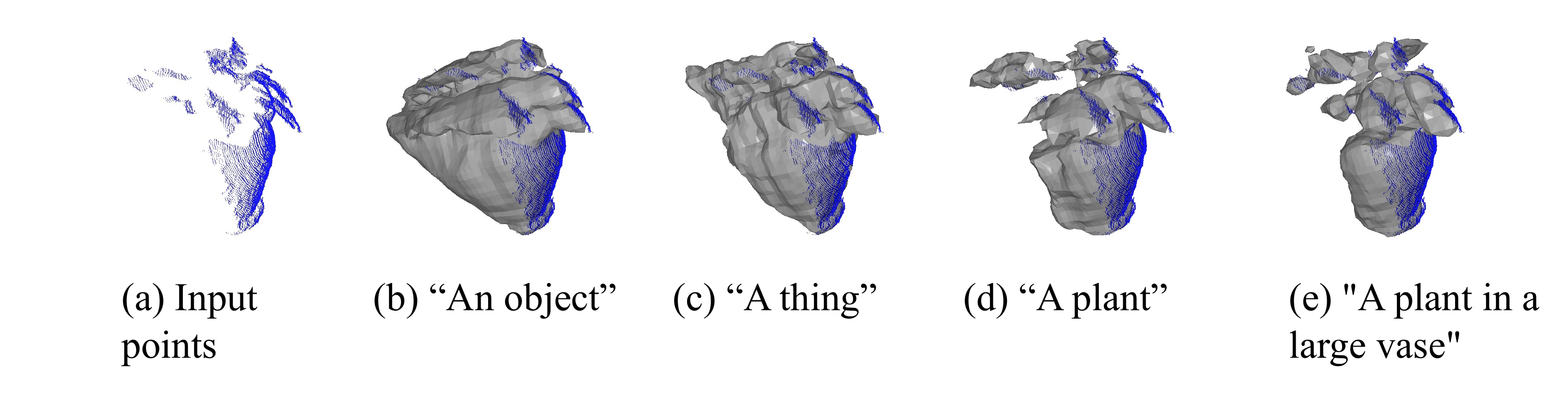}
    \caption{\textbf{The effect of using a specific or generic text description.} Results are shown for  reconstructing scan "06127" from the Redwood dataset. \textbf{(a)} The input point cloud. \textbf{(b, c)} Completion using two generic texts. Completion quality is poor. \textbf{(d)} Completion using the object class name. \textbf{(e)} Completion using a detailed textual description.} 
    \label{fig:fig_text}
\end{figure}

\begin{table}[h]
    \centering
     \begin{tabular}{l cccc}
         \toprule
         \midrule
         Object & "An object"  & "A thing" & "A $<$class name$>$" & Full text \\    
         \midrule
         executive chair & 32.9& 33.7& 28.9 &  \textbf{23.7} \\
         trash can & 37.8& 44.0& 37.2 &  \textbf{36.4} \\
         old chair & 86.6& 87.6& 22.4 &  \textbf{19.3} \\
         outside chair & 58.0& 52.7& 23.4 &  \textbf{22.6} \\
         plant in vase & 63.6& 44.5& 36.4 &  \textbf{29.5} \\
         one leg table & 48.3& 34.2& 24.2 &  \textbf{20.3} \\
         vespa & 70.6& 74.5& \textbf{57.6} &  \textbf{57.6} \\
         tricycle & 40.3& 44.7& \textbf{36.0} &  39.0 \\
         couch & 74.7& 69.5& \textbf{36.5} &  \textbf{36.5} \\
         office trash & 26.1& 26.7& 23.9 &  \textbf{20.5} \\
        \midrule
        Average & 53.9& 51.2&  32.6 &30.5 \\
    \bottomrule
    \end{tabular}
    \caption{\textbf{The effect of generic vs detailed prompt in terms of Chamfer distances (lower is better)}. Columns 1 and 2: two generic configurations where a global text is used for all objects. Column 3: only the class name is used e.g. both "executive chair" and "outside chair" are reconstructed with the text "A chair". Column 4: the results of our model with the text prompts from Table~\ref{tab:text_prompts}.      \label{tab:text_importance} }
\end{table}

\begin{SCfigure}
    \scalebox{1.01}{
    \setlength{\tabcolsep}{2pt} %
    \includegraphics[width=0.50\textwidth]{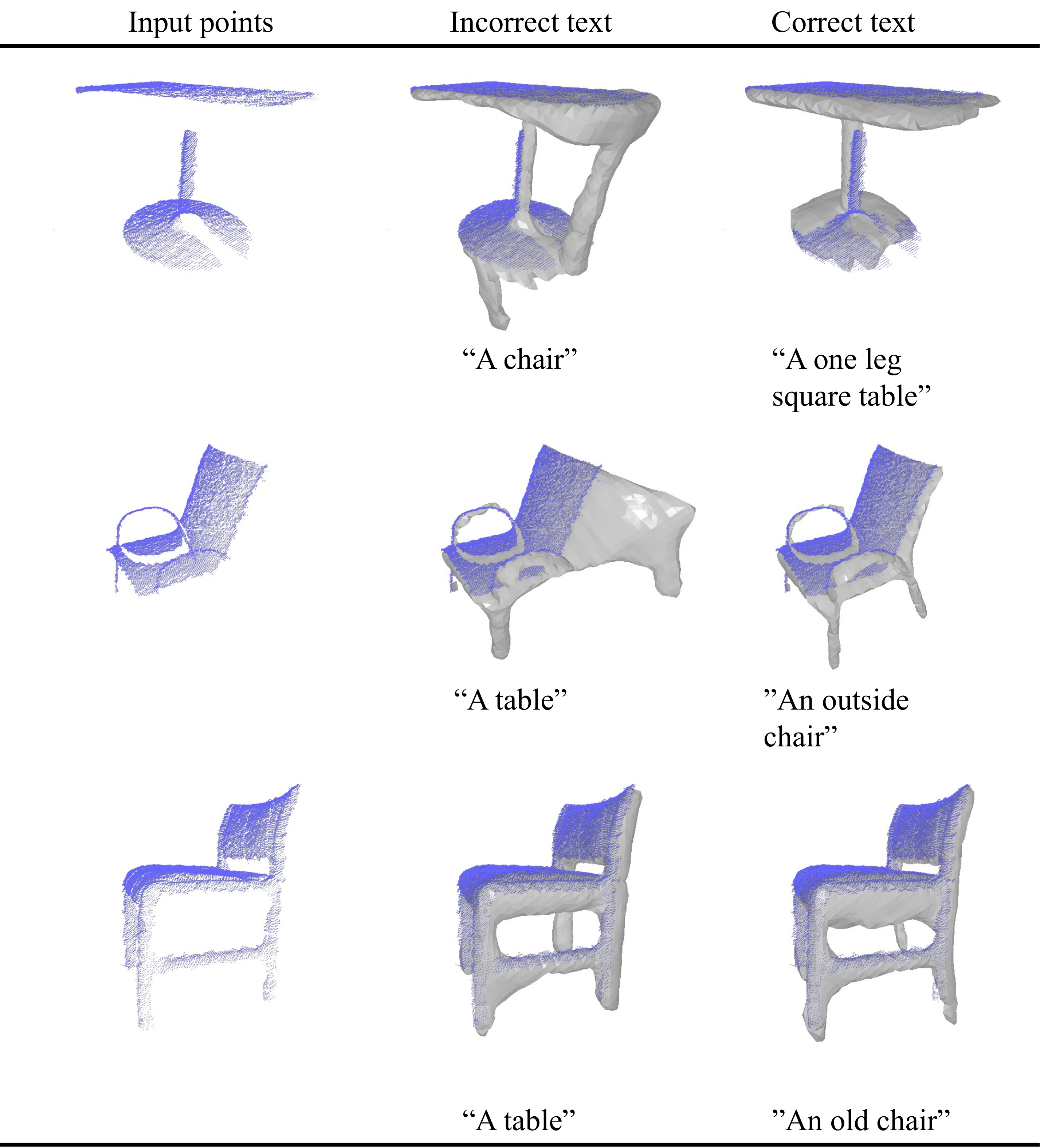}
    }
    \label{fig:fig_wrong_text}   
    \caption{The effect of incorrect  textual descriptions. Each row corresponds to a different object. \textbf{Left:} The partial scans that are given as input to our model. \textbf{Middle:} Completion performed using incorrect text descriptions. \textbf{Right:} the completion results of our method with our final text prompts. In the first two rows, the completion is inferior when given  the wrong text. In the bottom row, even with an incorrect text ("A table") the model still completes the chair correctly. This is because the input provides strong constraints. To make the shape more similar to a table, the method still needs to reconstruct the missing leg.    \label{fig:fig_wrong_text}}  
\end{SCfigure}

\subsection{Generic vs Detailed Text Prompts}
To evaluate the contribution of selecting an appropriate text prompt per object, we repeated reconstruction experiments of 
the $10$ objects  evaluated in the main paper, but varied the text prompts. 
Specifically, we used three levels of description specificity.  First, for a fully generic prompt (class agnostic), we tested two alternatives: "An object", "A thing". Second, we used the class name as the prompt. Finally, we used a more detailed description.

Table~\ref{tab:text_importance} provides the Chamfer distances between our reconstruction and the ground truth for all prompts. Using generic text yields inferior reconstructions. Adding specific details did not provide a significant improvement over using the class name. A qualitative comparison is shown in Fig.~\ref{fig:fig_text}. 


\subsection{Reconstruction with incorrect prompts}
We further check the sensitivity of our model to wrong text prompts. Specifically, we used the text: "A table" for a chair, and the text "A chair" for a table. The visualizations are presented in Fig.~\ref{fig:fig_wrong_text}.

 \begin{figure}[h!]
     \centering     \includegraphics[width=1.0\textwidth]{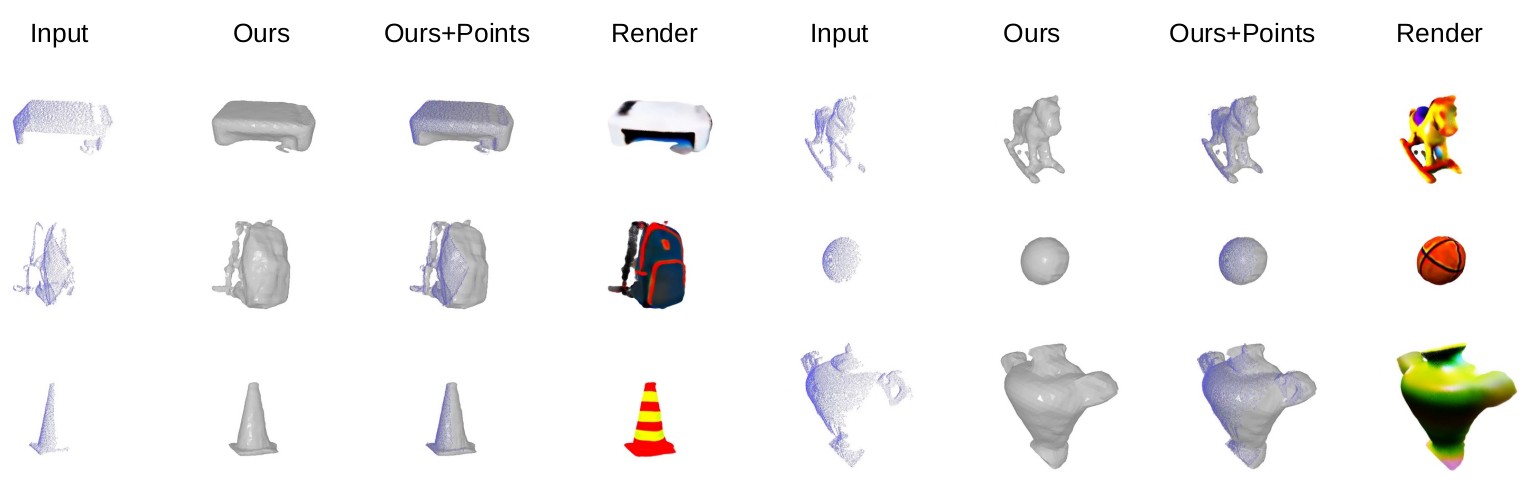}
     \includegraphics[width=1.0\textwidth]{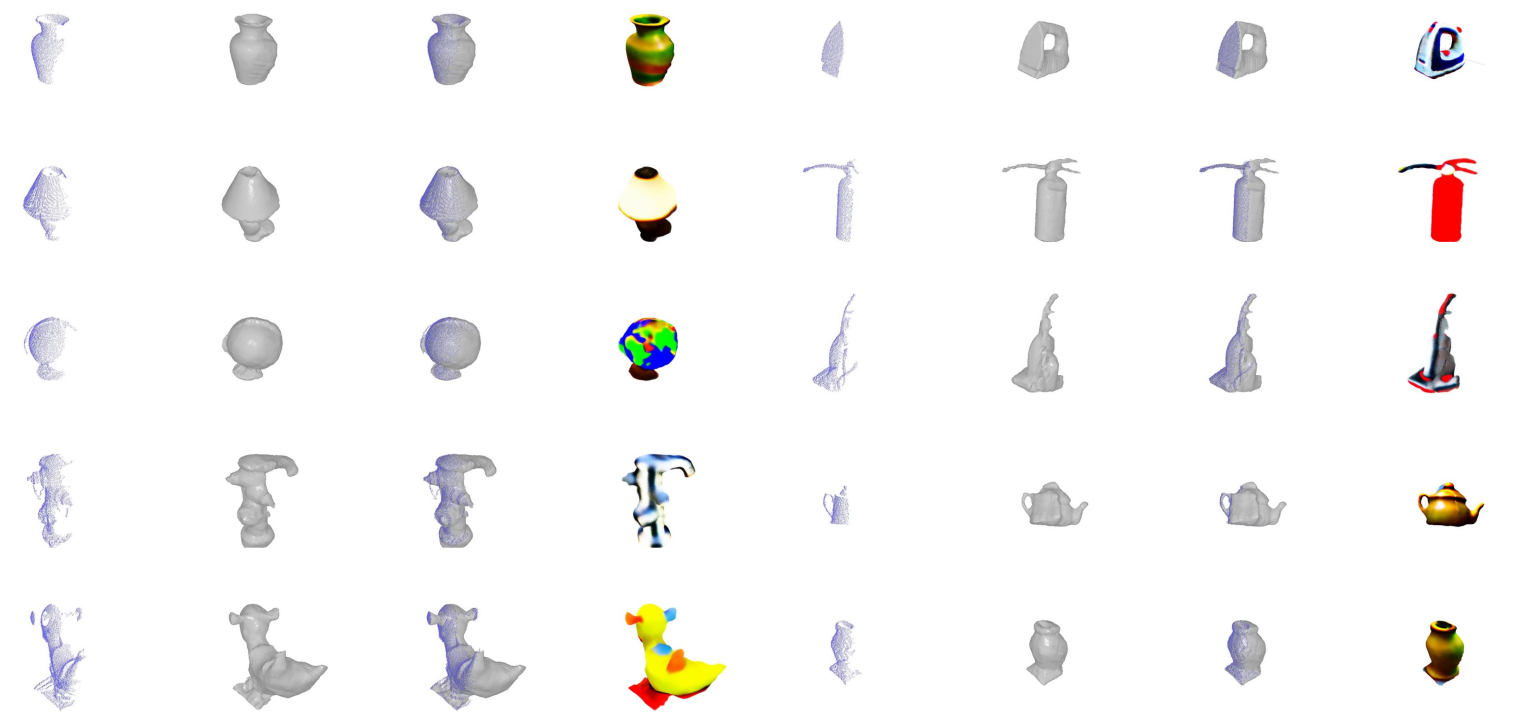}
     \caption{ Qualitative outputs of our method, when applied on Redwood cases with no available $360^o$ GT scans for quantitative comparison.  The figure is arranged as 2 columns of different objects, where for each object we show (from left to right): the input point cloud, the completed surface, the completed surface together with the input points, and image rendering of our optimized coloring function $\vv{c}$.}
     \label{fig:redwood_wo}
 \end{figure}

\section{Additional Results}
\begin{figure}[h!]
     \centering
     \includegraphics[width=\textwidth]{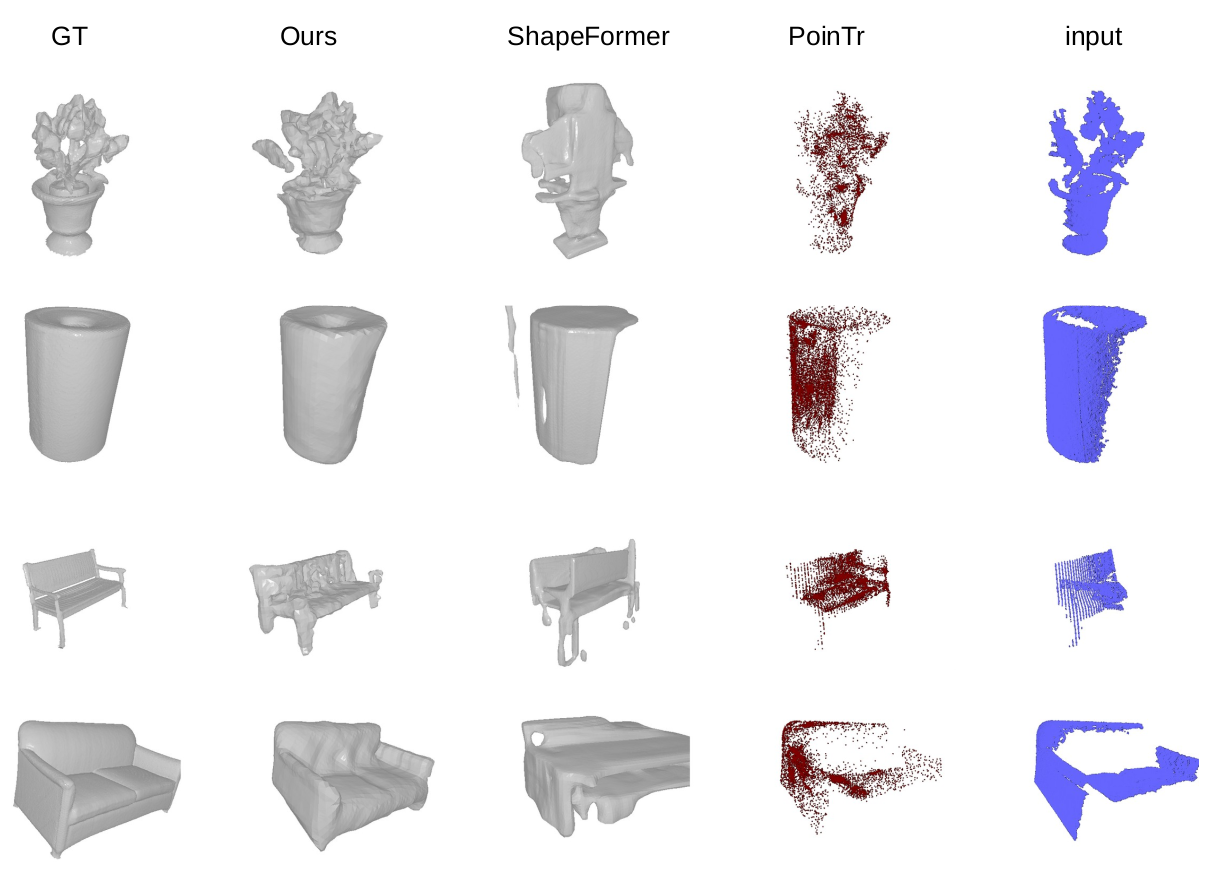}
     \caption{\textbf{Additional comparisons with four objects from the Redwood dataset.}  Qualitative comparisons between \ourmethod{} to baseline multi-class methods. Notably, \ourmethod{}, produces more accurate completions.} \label{fig:redwood_gt_supp}
\end{figure}

\vspace{-10pt}
\paragraph{Supplementary Redwood Results}
We evaluate our method on 4 additional Redwood cases with available ground truth surfaces: "08712","05456","00034" and "06912". Qualitative and quantitative comparisons are presented in Fig.\ref{fig:redwood_gt_supp} and Table~\ref{tab::redwood_table_gt} respectively. We can see that our method completes the shapes better than the baselines.

We applied our method to additional Redwood cases of various object types with no available ground truth. Qualitative results, including RGB renderings, are shown in Fig.~\ref{fig:redwood_wo}.
\paragraph{Video Results} We attach to the supplementary folder, $360^o$ video visualizations of our reconstructed objects for both, KITTI and Redwood datasets.

\begin{SCtable}
    \scalebox{1.01}{
    \setlength{\tabcolsep}{3pt} %
        \begin{tabular}{l rrr}
         \toprule
         \midrule
         object & Shape  & PoinTr & \ourmethod{} \\
          &  Former &  &  (ours) \\
         \midrule
         plant in vase 2 & 31.3 & 37.6 & \textbf{21} \\
         park trash can & 130 & 119.9 & \textbf{39.8} \\
         bench & 29 & 32.6 & \textbf{27.4} \\
         sofa & 106.6 & 129.3 & \textbf{29.2} \\
         \midrule
         \quad average & 82 & 83.9 & \textbf{33.7}\\
         \bottomrule
     \end{tabular}
       }
         \caption{\textbf{Quantitative evaluation for additional objects from the Redwood dataset.} Chamfer distance (in mm) comparisons between SDS-Complete to the baseline methods. Our method performs better than the baselines.}
    \label{tab::redwood_table_gt}
\end{SCtable}

\section{KITTI User Study.}
We conduct a user study for evaluating the various methods on the KITTI dataset given $15$ real object scans of cars, motorcycles, trucks, and an excavator. Specifically, we gather a group of 5 participants to rank the quality of each completed surface and its faithfulness to the input partial point cloud. For each object, the participants were given three anonymous shapes produced by the three methods: \ourmethod{}, ShapeFormer \cite{yan2022shapeformer}, and PoinTr \cite{yu2021pointr}. While the outputs of \ourmethod{} and ShapeFormer are surfaces, PoinTr only outputs a point cloud. Therefore, we applied Screened Poisson Surface Reconstruction \cite{kazhdan2013screened} to each output of PoinTr to base the user study comparisons on surface representations. The participants were instructed to choose the best shape, while the order of the methods was shuffled for each object. The best completion method for each input case is selected by the majority vote. Among the three methods, our method stands out as the method with the highest number of wins ($11$ out of $15$). 
 The results of the user study are presented in Table \ref{tab:user_study}.

\begin{table}[h!]
    \centering
    \begin{tabular}{l rrr}
         \toprule
         \midrule
         object & Shape  & PoinTr & \ourmethod{} \\
          &  Former &  &  (ours) \\
         \midrule
         best quality $[\%]$ & 26.6 & 0.0 & \textbf{73.4} \\
         \bottomrule
    \end{tabular}
    \vspace{10pt}
    \caption{\textbf{Human evaluation on KITTI}. A user study for evaluating our method's surface completion quality compared to the baselines, when tested on real object scans from the KITTI dataset. For each method, we present the percentage of cases that this method was selected by the participants as the best method. Our method gets the highest number of wins compared to the baselines. }
    \label{tab:user_study}
\end{table}



\section{Failure examples} 
 \begin{figure}[h!]
    \centering
    \includegraphics[width=0.95\textwidth]{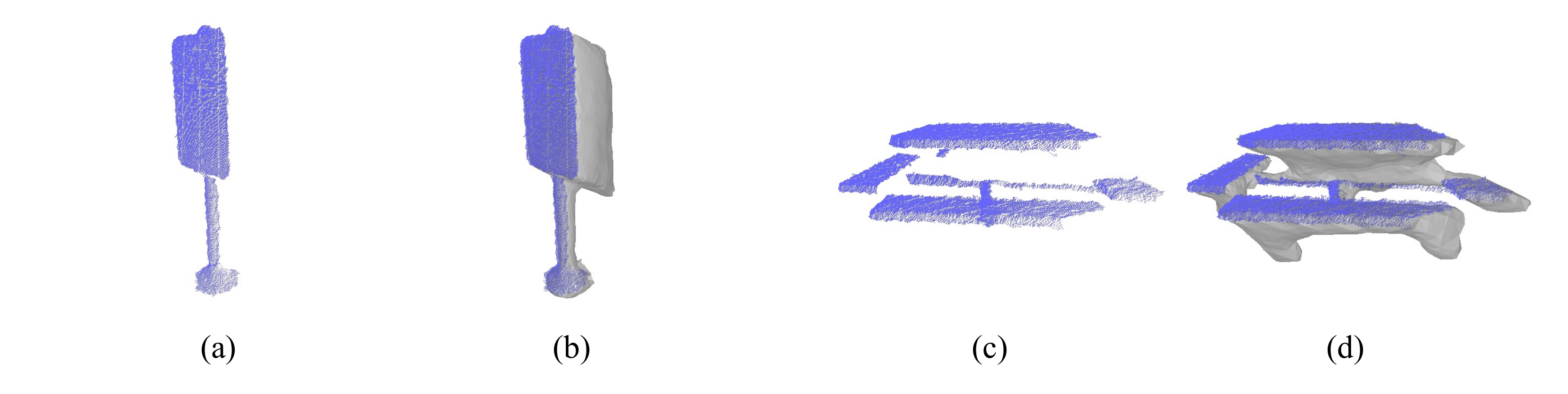}    
    \caption{\textbf{Failure cases of our method.} (a),(b) Input points and surface completion respectively, for Redwood scan "05492" (standing sign). (c),(d) Input points and surface completion respectively for Redwood scan "01373" (picnic table).    \label{fig:fig_failures}     
    }
\end{figure}
Fig.~\ref{fig:fig_failures} shows failure examples. In general, our method does not reconstruct well thin surfaces. We hypothesize that the initialization of the SDF to sphere \cite{Atzmon_2020_CVPR}, prevents the model from minimizing the occluded part at early training stages. Then, the SDS loss usually tries to paint this redundant content according to the text prompt, instead of removing it. Different initializations to the SDF, or other regularizations, need to be explored and left as future work.

\section{Implementation Details}


\paragraph{Running Time} We run our method for 2000 epochs, where each epoch uses 100 iterations. That takes about 1950 and 1380 minutes for Redwood and KITTI scans respectively, on NVIDIA RTX A6000. We note that many scans need much fewer iterations for converging, but to complete the fine details, e.g. the chair's legs, many iterations are needed due to the low resolution of the image rendering for the SDS-loss.

\paragraph{Network Architecture}For our optimized coloring function $\vv{c}_\varphi$, we use $4$ linear layers with $96$ channels, where the two intermediate ones are incorporated in ResNet \cite{he2016deep} blocks, as implemented by \cite{stable-dreamfusion}, with SiLU activations \cite{hendrycks2016gaussian}. For the SDF network ${f}_\theta$, we use 4 linear layers with $96$ channels and ReLU activations. $f_\theta$ is initialized to a sphere \cite{Atzmon_2020_CVPR} with radius lengths of $0.5$ and $0.9$ for Redwood and KITTI scans respectively. For both, $\vv{c}_\varphi$ and $f_\theta$ we use Positional Encoding with $6$ levels \cite{mildenhall2020nerf}. For extracting density from $f_\theta$ (Equations (5) and (6) in the main paper) we use $\alpha=100, \beta=10^{-3}$.

\paragraph{SDS-loss Implementation Details} We base our code on the implementation of \cite{stable-dreamfusion}. During training, for each iteration, we randomly sample a background color, to prevent the model from "changing" the geometry by just coloring it with the background color. For Redwood cases, we render $80\times 80$ images for the SDS-loss using the sampled camera and the known internal parameters of the sensor. For KITTI, at initialization time we first project the object's LiDAR points to a 2D spherical projection \cite{behley2019iccv}, with height and width of $64$ and $1024$ pixels respectively. We use the projected 2D mask to select the 2D bounding box area of $64$ pixels height, where the width is determined by the min and max horizontal coordinates of the object $\pm 5$ pixels. The LiDAR rays that define this selected bounding box are used to render the object during training, where a novel camera pose is defined by rotating these rays around the object's centroid.  As a text-to-image diffusion model we use  Stable Diffusion v2  \cite{rombach2022high}.
\paragraph{Training Details} We optimize the networks using the Adam optimizer \cite{kingma2014adam} with a learning rate $10^{-4}$. The coefficients for our loss for all the experiments are $\delta_{m}=10^5$, $\delta_{d}=10^5$, $\delta_{p}=10^5$, $\delta_{eikonal}=10^4,\delta_{plane}=10^5$. At each iteration we sample $1000$ uniform points for $\mathcal{L}_{plane}$ and $\mathcal{L}_{eikonal}$. For $\mathcal{L}_m,\mathcal{L}_d$, at each iteration,  we randomly sample $2000$ pixels for Redwood cases, whereas for KITTI, we render the entire bounding box.
\paragraph{Camera Sampling} As  described in Section 4.2 of the main paper, during training, we start by applying the SDS loss on the rendered image from $C_0$ pose, and then we gradually increase the sampling range of the deviation angles until the entire object is covered. 
In more detail, we gradually increase the sampling range of the azimuth angles: $\gamma_\text{azimuth}\sim\mathcal{U}(-\nu,\nu) $, starting from $\nu=0$ to $\nu=180$. Specifically, we set $\nu=30,45,60,90,180$ at epochs $20,50,80,100,120$ respectively.   $\gamma_\text{elevation}$ is set to $0$ for $20$ epochs and then uniformly sampled according to: $\gamma_\text{elevation}\sim\mathcal{U}(-\xi_0,0)$ for Redwood scans, where $\xi_0$ is the elevation of $C_0$ from the plane $\vv{l}$ in degrees. For KITTI scans (after epoch 20)  we use $\gamma_\text{elevation}\sim\mathcal{U}(-\xi_0,\xi_0)$ since the original viewpoint is usually low, and we also scale the distance from the source to the object uniformly by $\sim\mathcal{U}(1,2)$ after epoch $20$. As in \cite{poole2022dreamfusion}, we augment the input text according to the viewing direction, with a text that describes the viewpoint orientation. Specifically, as in \cite{stable-dreamfusion} we use "*, front view", "*, side view", "*, back view",  "*, overhead view" and "*, bottom view", where * denotes the input text. Unlike \cite{poole2022dreamfusion}, the orientation of the object is determined by the input points. Therefore, we use an extra input from the user of $\gamma_{0_\text{azimuth}}$, which explains the original viewpoint, e.g. $\gamma_{0_\text{azimuth}}=90$ if the object is viewed from the side. Then, during training, we use $\gamma_{0_\text{azimuth}}$ and $\gamma_\text{azimuth}$ to calculate the azimuth with respect to the object, and $\gamma_\text{elevation}$ to compute the elevation with respect to the plane $\vv{l}$. These orientations are used to augment the text with the corresponding view direction description.
\paragraph{Object Centralization} Given the input points we centralize them at the origin. This is done in general by subtracting their center of mass. When the object's largest dimension is aligned with the viewing axis, the center of mass is usually biased toward the camera. To handle this, we extract an oriented 3D bounding box for the input points and measure the ratio between the largest distance to the smallest distance from the center of mass to any bounding box point. If this ratio is above $1.7$ we use the bounding box center as our centroid instead of using the center of mass. In the KITTI dataset, which mostly includes non-isotropic objects,  we always use the   bounding box center as our centroid. We then scale the points such that the largest point norm is $0.5$.   
\paragraph{Baseline Runnings} For running the baseline methods, we tried to locate the input points as much as possible according to the method's expectations to prevent them from failing. This includes using our knowledge about the world plane $\vv{l}$ and the object orientation with respect to the camera $\gamma_{0_\text{azimuth}}$. For ShapeFormer, each time we took the best shape out of the 5 that it outputs.
\paragraph{Data Processing} 
For the Redwood dataset, we segmented out the foreground object manually.  As a preprocessing, we manually aligned the GT scan with the partial point cloud and applied ICP for refinement \cite{Arun1987LeastSquaresFO}. 
Each KITTI scan that we used, is the aggregation of 5 timestamps. The segmentation map for KITTI is given by \cite{behley2019iccv}. For both, KITTI and Redwood datasets and for each scan,  the plane $\vv{l}$ is segmented out from the original point cloud using RANSAC \cite{hartley2003multiple}.

\section{List of Textual Prompts}
The text prompts for the Redwood shapes that we tested are presented in Table~\ref{tab:text_prompts}. For the KITTI dataset, we used the text "A *", where "*" denotes the shape class: "car", "truck", "motorcycle", or "mini excavator".
\begin{table}[h!]

    \centering
    \begin{tabular}{ c l l }
 Scan ID & Scan name & Text Prompt \\   
\midrule
 08712 & plant in vase 2& "A plant" \\
 06912 & bench& "A bench" \\
 05492 & standing sign& "A standing sign" \\
 05456 & park trash can& "Park trash can" \\
 01373 & picnic table& "A picnic table" \\
 00034 & sofa& "A sofa" \\
 01184 & trash can& "An outdoor trash can with wheels" \\
 06127 & plant in vase& "A plant in a large vase" \\
 06830 & tricycle& "Children's tricycle with adult's handle" \\
 07306 & office trash& "An office trash can" \\
 05452 & outside chair& "An a outside chair" \\
 06145 & one leg table& "A one leg square table" \\
 05117 & old chair& "An old chair" \\
 09639 & executive chair& "An executive chair" \\
 06188 & vespa& "A motorcycle" \\
 07136 & couch& "A couch" \\
 08754 & teapot& "A teapot" \\
 09664 & traffic cone& "A traffic cone" \\
 04797 & street faucet& "A street faucet" \\
 02426 & clothes iron& "A clothes iron" \\
 09424 & backpack& "A backpack" \\
 04014 & lamp& "A lamp" \\
 04210 & globe& "A globe" \\
 06271 & vase& "A vase" \\
 04154 & fire extinguisher& "A fire extinguisher" \\
 00030 & vase 2& "A vase" \\
 09484 & rocking horse& "A rocking horse" \\
 04919 & duck spring swing& "A duck spring swing" \\
 06177 & printer& "A printer" \\
 04457 & vacuum cleaner& "Vacuum cleaner" \\
 06124 & vase 3& "A vase" \\
 09488 & basketball& "A basketball" \\
 \midrule
 \bottomrule
    \end{tabular}
    \caption{Scan IDs from the Redwood dataset \cite{choi2016large}, and their corresponding textual prompts.}
    \label{tab:text_prompts}
\end{table}

\section{Broader Impact}
Our approach  uses the SDS loss that builds on a text-to-image diffusion model  \cite{rombach2022high}. As such, it inherits possible biases that such a model may have. 

\end{document}